\documentclass[runningheads]{llncs}
\usepackage{times}
\usepackage{soul}
\usepackage{url}
\usepackage[hidelinks]{hyperref}
\usepackage[utf8]{inputenc}
\usepackage[small]{caption}
\usepackage{graphicx}
\usepackage{amsmath}

\begin{document}
\title{Improving Transferability of Deep Neural Networks}

\author{Parijat Dube\inst{1}
\and
Bishwaranjan Bhattacharjee\inst{1}
\and
Elisabeth Petit-Bois\inst{1,2}\thanks{Work done when the author was a student intern at IBM Research AI.}
\and
Matthew Hill\inst{1}
}
\authorrunning{P. Dube et al.}
%
\institute{IBM Research AI, Yorktown Heights NY 10598, USA\\
\email{pdube,bhatta,mh@us.ibm.com}
\and
Kennesaw State University, Kennesaw GA, 30144, USA\\
\email{epetitbo@students.kennesaw.edu}}
\maketitle             

\begin{abstract}
Learning from small amounts of labeled data is a challenge in the area of deep learning. This is currently addressed by Transfer Learning where one learns the small data set as a transfer task from a larger source dataset. Transfer Learning can deliver higher accuracy if the hyperparameters  and source dataset are chosen well. One of the important parameters is the learning rate  for the layers of the neural network.  We show through experiments on the ImageNet22k and Oxford Flowers datasets that improvements in accuracy in range of  127\% can be obtained by proper choice of learning rates.  We also show that the images/label parameter for a dataset can potentially be used to determine optimal learning rates for the layers to get the best overall accuracy. We additionally validate this method on a sample of real-world image classification tasks from a public visual recognition API.

\end{abstract}

\section{Introduction}
Deep Learning has become all pervasive in many application domains like Vision, Speech, and Natural Language Processing \cite{nature}. This can be partly attributed to the availability of fast processing units like GPUs as well as better neural network designs.   The availability of large, open source, general purpose labeled data has also helped the penetration of Deep Learning  into these domains.

\begin{figure}[ht]
   \includegraphics[width=\textwidth]{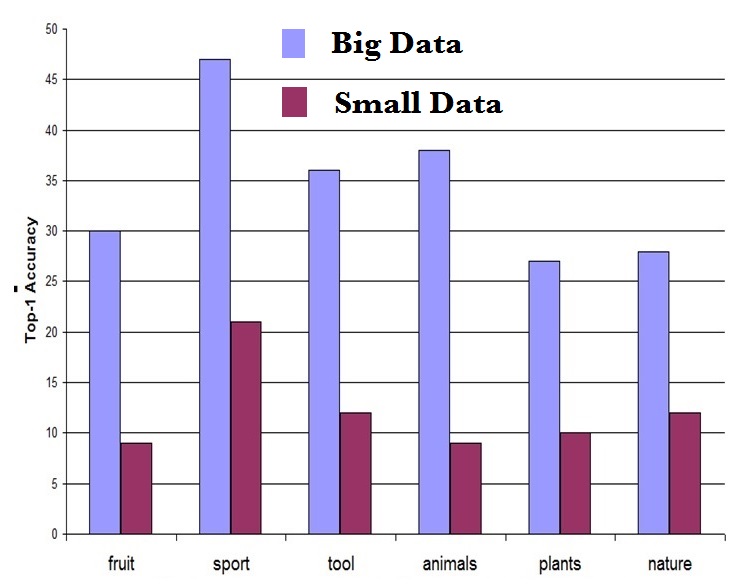}
   \caption{Impact of data size on learning accuracy}
   \label{fig.bigdata}
\end{figure}

The accuracy obtained on a learning task depends on the quality and quantity of training data. As Figure \ref {fig.bigdata} shows, with larger amounts of data, for the same learning task,  one can obtain much better accuracy.  In this figure, the accuracy obtained on various categories of  ImageNet22K \cite {imagenet22k} are shown with the big data being 10x bigger in size than the small data. While large, open source, general purpose, labeled data is available,  customers often have specific  needs for training.  For example, a doctor may be interested in using Deep Learning for Melanoma Detection \cite {Codella:2015}.  The amount of labeled data available in these specific areas is rather limited.  In situations like these,  the training accuracy can be negatively impacted if trained with only this limited data. To alleviate this problem, one can fallback on Transfer Learning \cite {pan:TKDE}  \cite {mou2016}.  

In Transfer Learning, one takes a model, trained on a potentially large dataset (called the source dataset) and then learns a new, smaller dataset (called the target dataset) as a transfer task (T) on it.  This can be achieved by finetuning the weights of neurons in the pre-trained model using the target dataset. Finetuning is a technique to leverage the information contained in a source dataset by tweaking the weights of its pre-trained network while training the model for a target  dataset. It has been shown that models trained on the source dataset learn basic concepts which will be useful in learning the target dataset \cite {Yosinski:nips1}.

In the area of vision, the neural networks tend to be quite deep in terms of layers \cite {ResNet27}. It has been shown that the layers learn different concepts. The initial layers learn very basic concepts like color, edges, shapes, and textures while later layers learn complex concepts \cite {alex}. The last layer tends to learn to differentiate between the labels supported by the source dataset. 

\begin{figure}[ht]
   \includegraphics[width=\textwidth]{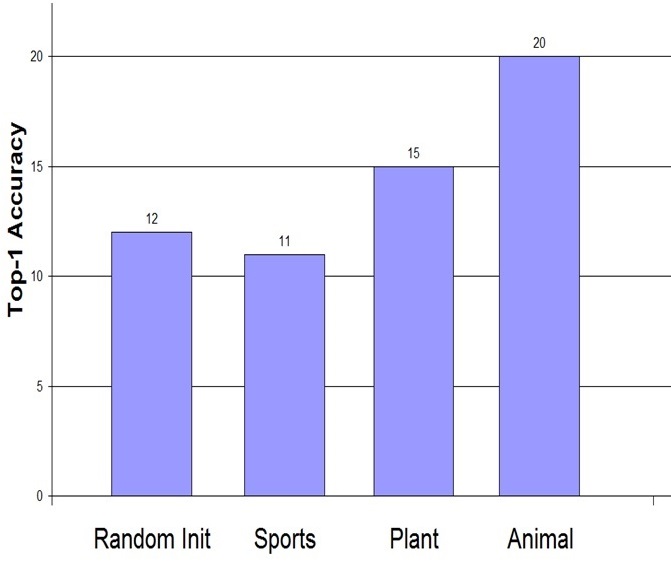}
   \caption{Impact of base model on transfer learning accuracy}
   \label{fig.basemodel}
\end{figure}

The key challenges to Transfer Learning are  how, what and when to transfer \cite {pan:TKDE}. One needs to address key  questions like the selection of the source dataset, the neural network to use, the various hyperparameter settings as well as the type of training method to apply on the selected neural network and dataset.  Figure  \ref {fig.basemodel} shows the accuracy obtained while training on the Tool category of ImageNet22K on models created from different source categories of ImageNet22K like Sports, Animals, Plant as well as random initialization. As the figure indicates, accuracy varied from -8\% to +67\% improvement  over the random initialization (no Transfer Learning) case.

When performing Transfer Learning using deep learning, a popular method of training is using Stochastic Gradient Descent (SGD) \cite {SGD}. In SGD, the key hyperparameters to control the descent are the block size, the step size and the learning rate. In the case of Transfer Learning, the learning rate can be set for every layer of the neural network. This controls how much the weights in each layer change as training progresses on the target dataset.  A lower learning rate for a layer allows the layer to retain what it has learned from the source data longer. Conversely, a higher learning rate forces the layer to relearn those weights quicker for the target dataset. For Transfer Learning, the concepts learned in the early layers tend to have high value since the source dataset is typically large, and the early layers represent lower-level features that are transferable to the target task. 
If the rates are large, then the weights could change significantly  and the neural network could over-learn on the target task, especially if the target task has a limited amount of training data.  The accuracy that is obtained on the target task depends on the proper selection of all these parameters.

In this paper we study the impact of individualized layer learning rates on the accuracy of training. We use a  large dataset called ImageNet22K \cite {imagenet22k} and a small dataset called the Oxford Flowers \cite {oflower} for our experiments.  These experiments are done on a deep residual network \cite {ResNet27}. We show that the number of images-per-label plays an important role in the choice of the learning rate for a layer. We also share preliminary results on real world image classification tasks which indicate graduated learning rates across a network, such that early layers change slowly, allow for better accuracy on the target dataset. 

The paper is organized as follows: In section 2,  we describe related work. In sections 3 and 4 we describe our experimental setup and present result our results, respectively. We conclude in section 5.

 \begin{figure}[ht]
   \includegraphics[width=\textwidth]{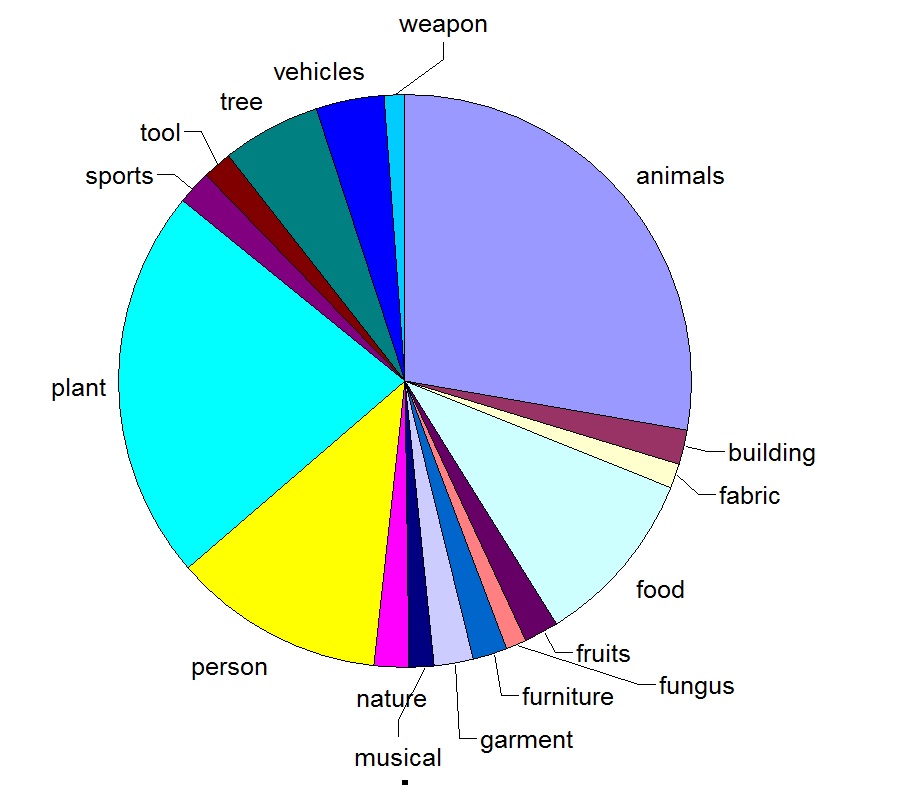}
   \caption{Imagenet22k hierarchies used}
   \label{fig.imagenet22k}
\end{figure}

\section{Related Work}
Several approaches are proposed to deal with the problem of learning with small amounts of data. These include one-shot learning~\cite{Fei-Fei:2006}, zero-shot learning~\cite{Palatucci:2009}, multi-task learning~\cite{Argyriou:2006}\cite{Dong2015MultiTaskLF}, and generic transfer learning~\cite{Yosinski:nips1} \cite{transfer1} \cite{Bhattacharjee2017}. 

Multi-task learning simultaneously trains the network for multiple related tasks by finding a shared feature space \cite{Argyriou:2006}. An example is Neural Machine Translation (NMT) where the same network is used for translation to different languages \cite{Dong2015MultiTaskLF}.  
In \cite{transfer1} a joint fine-tuning approach is proposed to tackle the problem of training with insufficient labeled data. The basic idea is to select a subset of training data from source dataset (with similar low-level features as target dataset) and use it to augment the training dataset for target task. Here the convolutional layers of the resulting network are finetuned for both the source and target tasks. Our work is targeted for scenarios where source dataset is not accessible and finetuning is only possible using a target dataset.  

 It was established in \cite {Yosinski:nips1} that finetuning all the layers of the neural network gives the best accuracy. However there is no study on the sensitivity of accuracy to the degree of finetuning. In \cite {Bhattacharjee2017} it is experimentally shown for one dataset that the accuracy of a (finetuned) model monotonically increases with increasing learning rate and then decreases, indicating existence of an optimal learning rate before overlearning happens. We studied variation in accuracy of model with learning rate used in finetuning for several datasets and observed non-monotone patterns. 

Another popular form of Transfer Learning is by using deep feature embeddings from a neural network to drive binary Support Vector Machines (SVMs) \cite {Donahue:2014} \cite{Bhattacharjee2017}. In this approach, there are as many SVMs as categories in the target dataset and each SVM learns to classify a particular label. The feature embeddings can be taken from any layer of the neural network but, in general, is taken from the penultimate layer. This is equivalent of fine tuning with the learning rate multipliers of all the inner layers up to the penultimate layer being kept to 0 and the last layer being changed.

\section{Experimental Setup}
\begin{figure}[ht]
   \includegraphics[width=\textwidth]{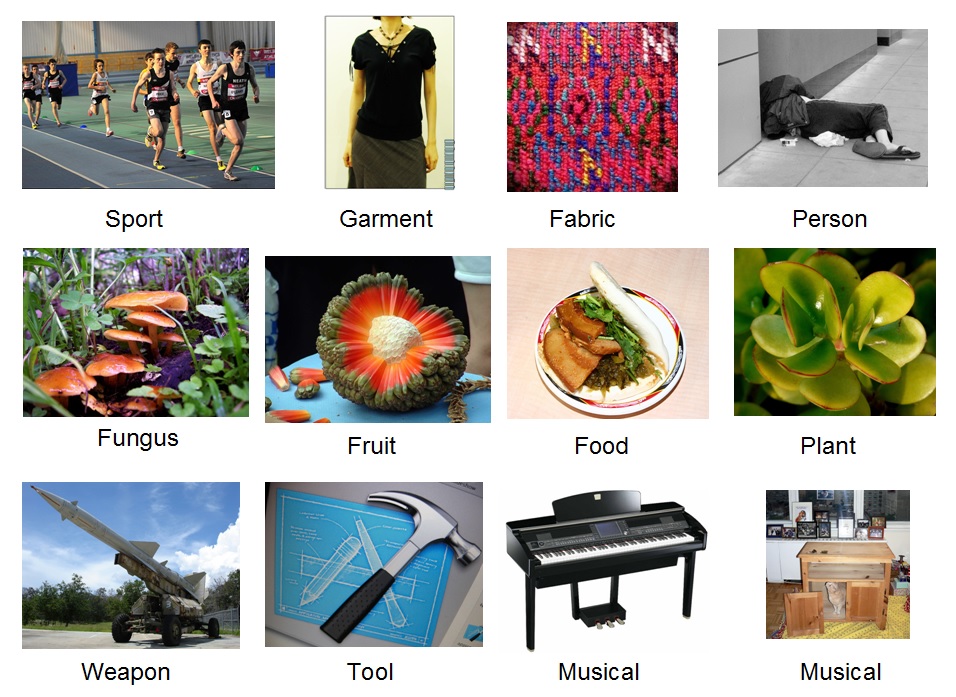}
   \caption{Representative images from various Imagenet22k  hierarchies used in experiments}
   \label{fig.22kimgs}
\end{figure}

ImageNet22k contains 21841 categories spread across hierarchical categories. We extracted some of the major hierarchies  like  sport, garment, fungus, weapon, plant, animal, furniture, food, person, nature, music, fruit, fabric, tool, and building to form multiple source and target domains image sets for our evaluation.   Figure \ref {fig.imagenet22k} shows the hierarchies of ImageNet22k dataset that was used and their relative sizes in terms of number of images. Figure \ref {fig.22kimgs} show representative images from some of these important domains. Some of the domains like animal, plant, person, and food contained substantially more images (and labels) than categories such as weapon, tool, or sport. This skew is reflective of real world situations and provides a natural testbed for our method when comparing training sets of different sizes.

Each of these domains was then split into four equal partitions. One was used to train the source model, two were used to validate the source and target models, and the last was used for the Transfer Learning task. One-tenth of the fourth partition was used to create a Transfer Learning target. For example, the person hierarchy has more than one million images. This was split into four equal partitions of more than 250K each. The source model was trained with data of that size, whereas the target model was fine-tuned with one-tenth of that data size taken from one of the partitions. The smaller target datasets are reflective of real Transfer Learning tasks.

We augmented the target datasets by also using the Oxford Flower dataset \cite {oflower} as a separate domain. The dataset contains 102 commonly occurring flower types with 8189 images. Out of this, a target dataset of only 10 training images per class was used. The rest of the data was used for validation. 

\begin{figure}[ht]
   \includegraphics[width=\textwidth]{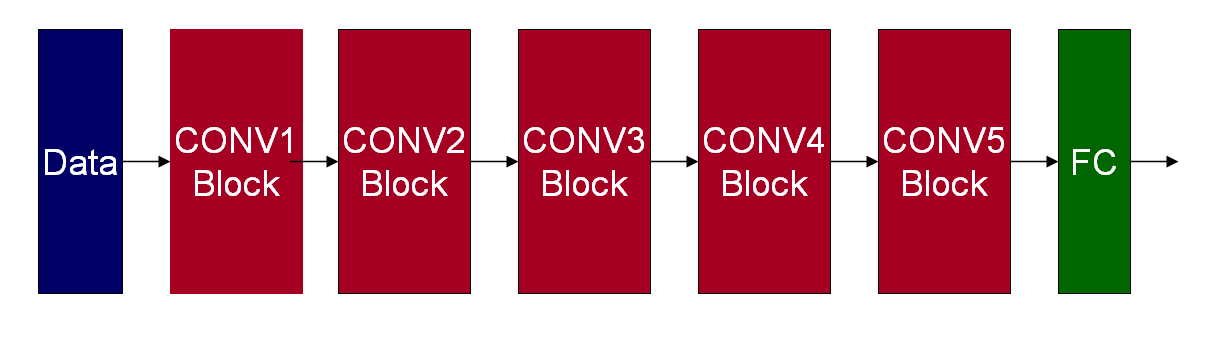}
   \caption{Major Blocks of the ResNet model used in the experiments}
   \label{fig.ResNet27}
\end{figure}

The training of the source and target models was done using Caffe \cite {caffe1} and a ResNet-27 model \cite {ResNet27}. The main components of this neural network  are shown in Figure \ref{fig.ResNet27}.  The source models were trained using SGD \cite {SGD} for 900,000 iterations with a step size of 300,000 iterations and an initial learning rate of 0.01. The target models were trained with an identical network architecture, but with a training method with one-tenth of both iterations and step size. A fixed random seed was used throughout all training. 

\section{Results and Discussion}
 Finetuning the weights involves initializing the weights to the values from the source model and then adjusting them to reduce the classification loss with the target dataset. Typically in fine-tuning a source model to a target domain, the practice is to keep the weights of all the inner layers unchanged and only finetune the weights of the last fully connected layer. The parameter which controls the degree of finetuning is the learning rate. Let $IL-n/LL-m$ be a transfer learning finetuning experiment where the inner layers learning rate ($IL$) is at $n$ and outer layer learning rate ($LL$) is at $m$, with $n<m$. We are assuming a uniform learning rate for all the inner layers for most of the experiments.  For those where the inner learning rate was varied, it is specifically mentioned in the paper. 
 \subsection{Finetuning Last Layer}
 We first did some experiments to quantify the gains possible by varying the learning rate of the last layer in finetuning while keeping all the inner layers weights unchanged. Table~\ref{tbl.IL0LL} compares the difference in accuracy of trained model for two different values of learning rate of the last layer, 0.01 and 0.1, corresponding to experiments $IL-0/LL-0.01$ and $IL-0/LL-0.1$. Observe that the accuracy is sensitive to the choice of $LL$ and significant gains in accuracy (up to 127\%) are achievable for certain domains by just choosing the best value of $LL$. 
 \begin{table}
     \centering
     \captionof{table}{Transfer Learning Accuracy with varying $LL$}\label{tbl.IL0LL}
     \begin{tabular}{rrrrr}
     \hline
     \bf Target& \bf Source & \bf LL-0.01 & \bf LL-0.1 & \bf \% Gain\\
     \hline
     fabric & garment & \bf 13.09\% & 11.33\% & 15.47\%\\
     tool & weapon & 14.54\% & \bf 14.78\% & 1.63\% \\
     oxford & plants & \bf 91.06\% & 73.17\% & 24.44\% \\
     food & fruit & \bf 5.71\% & 5.07\% & 12.52\% \\
     fungus & plant & \bf 13.12\% & 5.80\% & 127.79\% \\
     person & food & \bf 4.49\% & 2.81\% & 59.75\%\\
     fruit & garment & 9.30\% & \bf 10.50\% & 12.92\%\\
     music & plant & \bf 15.37\% & 9.47\% & 62.22\%\\
     \hline 
     \end{tabular}
\end{table}

\subsection{Finetuning Inner Layers} 
An earlier work \cite{Yosinski:nips1,Bhattacharjee2017} has observed that finetuning inner layers along with the last layer can give better accuracy compared to only finetuning the last layer. However their observation was based on limited datasets. 
We are interested in studying how the accuracy changes with $IL$ for a fixed $LL$ with following objectives: 
\begin{itemize}
   \item[(i)] Identify patterns which can be used to provide guidelines for choosing $LL$ and $IL$ for a give source/destination dataset.
   \item[(ii)]Find correlation between dataset features like images/label, similarity between source and target datasets, and the choice of $IL/LL$.
   \item[(iii)] Quantify possible gains in accuracy for different datasets by exploring the space of $LL$ and $IL$ values and hence establish the need to develop algorithms for identifying the right set of fine tuning parameters for a given source/target dataset. 
\end{itemize}

\begin{figure}[ht]
   \centering
   \includegraphics[width=3.3in]{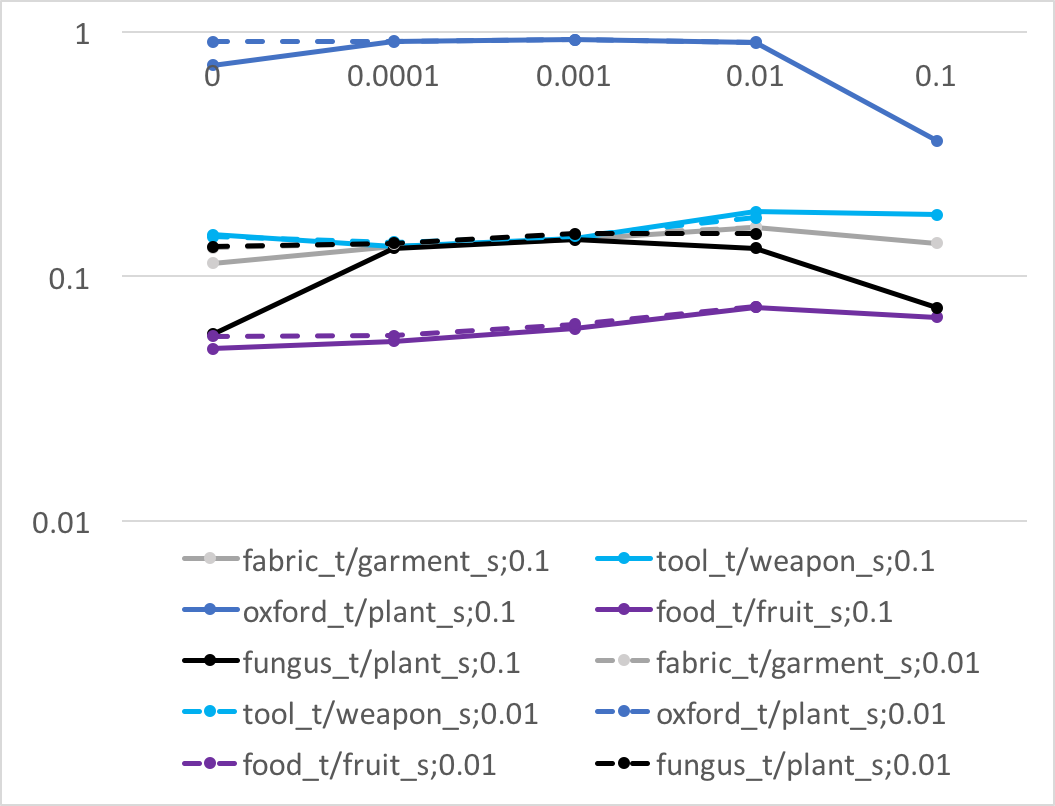}
   \caption{Set A accuracy vs IL for fixed LL}
   \label{fig.setAaccuracyVsIL}
\end{figure}

\begin{figure}[ht]
   \centering 
   \includegraphics[width=3.3in]{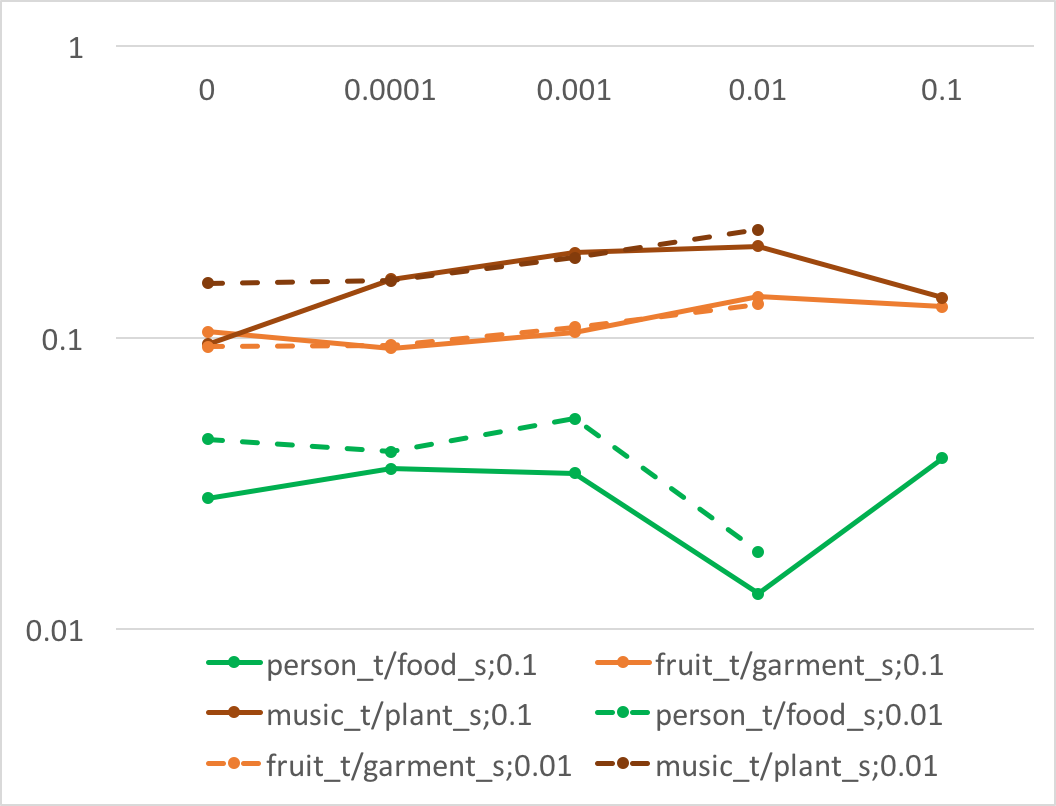}
   \caption{Set B accuracy vs IL for fixed LL}
   \label{fig.setBaccuracyVsIL}
\end{figure}

To this end, we conducted experiments varying $IL$ for a fixed $LL$. We divided the experiments into two sets based on perceived semantic closeness of source and target domains. Set A (B) consists of experiments where the source and target datasets are semantically close (far). Thus we have,
\begin{align*}
A = \{ & fabric_t/garment_s, tool_t/weapon_s,oxford_t/plants_s,\\
& food_t/fruit_s,fungus_t/plant_s \}, \text{and} \\
B=\{&person_t/food_s,fruit_t/garment_s,music_t/plant_s\}
\end{align*}
 Figures~\ref{fig.setAaccuracyVsIL} and \ref{fig.setBaccuracyVsIL} show the accuracy obtained by increasing $IL$ by powers of 10 between 0 and $LL$ for $LL=0.01$ and $0.1$. So when $LL=0.01 (0.1)$, $IL$ took values in $\{0,0.0001,0.001,0.01\}(\{0,0.0001,0.001,0.01,0.1\})$. 

\begin{figure}[ht]
   \includegraphics[width=\textwidth]{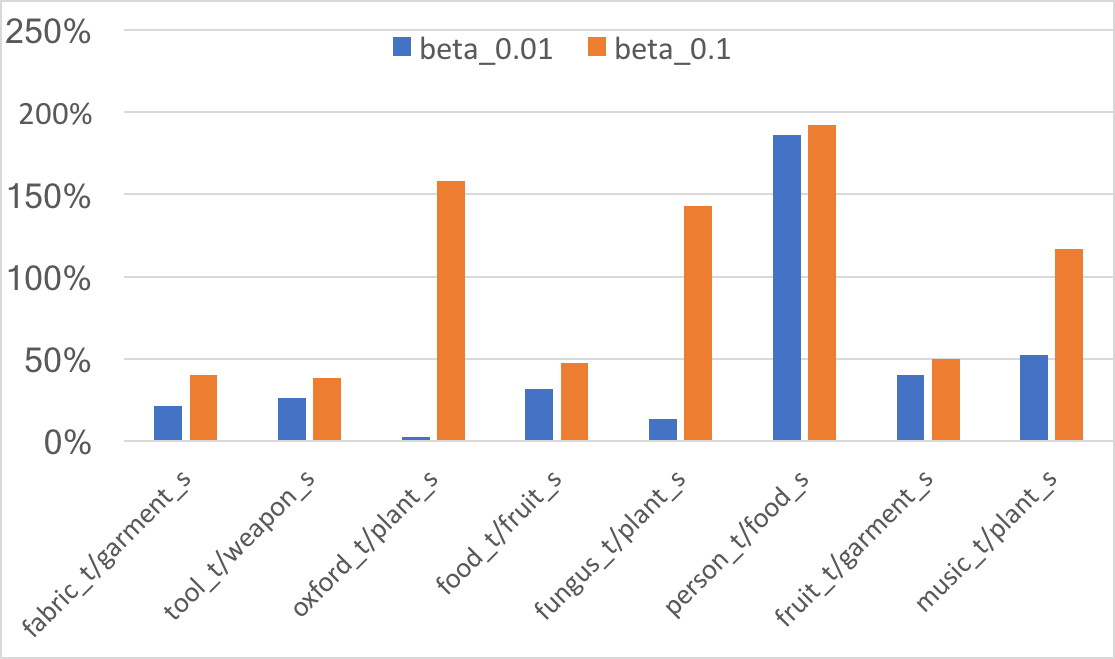}
   \caption{Range of variation in accuracy with varying $IL$}
   \label{fig.beta_0.01_0.1}
\end{figure}

Two patterns across different experiments are observed: (i) accuracy increases monotonically with $IL$ and then decreases (ii) accuracy alternates between increase and decrease cycles. The variation in accuracy with $IL$ can be significant for certain datasets. Let $min_m$ and $max_m$ be the minimum and maximum value of accuracy obtained when $IL$ is varied at $LL=$ and $\beta_m$ be defined as: 
\begin{equation}m
  \beta_m = \frac{max_m - min_m}{min_m}\times100  
\end{equation}
Observe that $\beta_m$ represents the percentage range of possible variation in accuracy with $LL=m$ and varying $IL$. Figure~\ref{fig.beta_0.01_0.1} compares $\beta_m$ for different datasets. All the datasets exhibit $\beta_m>0$, with median values of $\beta_{0.01} (\beta_{0.1})$ being 28.96\% (83.52\%). Observe that $\beta_{0.1}> \beta_{0.01}$ for all the datasets. 
Also, for same dataset, the range of variation in accuracy can be quite large or small depending on $LL$. For example, for $oxford_t/plant_s,fungus_t/plant_s$ and $music_t/plant_s$ the difference $\beta_{0.1}-\beta_{0.01}$ is greater than 100 points. 
Thus, finetuning both inner and outer layers gives the best accuracy. Further the value of $IL$ that maximizes accuracy can be different for different datasets. The pattern of variation in accuracy with $IL/LL$ is not always monotone. 

Let $\alpha_m$ be the value of $IL$ that achieves the best accuracy $(max_m)$ at $LL=m$ for a dataset.
Table~\ref{tbl.alpham} lists $\alpha_m$ for different datasets. The last column in the table shows the difference $max_{0.1}-max_{0.01}$. Observe that there is no clear winner, for some datasets keeping $LL=0.1$ and then searching for $IL$ gives the best accuracy while for others $LL=0.01$ performs better. This indicates the need for joint optimization over the space of $LL$ and $IL$ to get the best accuracy. 
\begin{table}[]
    \centering
    \captionof{table}{$\alpha_m$ for different datasets under study}\label{tbl.alpham}
    \begin{tabular}{rrrrr}
    \hline
    \bf Target& \bf Source & \bf $\alpha_{0.01}$ & \bf $\alpha_{0.1}$ & $\max_{0.1}-\max_{0.01}$ \\
    \hline
    fabric & garment & 0.0001& 0.0001&0.00\%\\
    tool & weapon & 0.0001& 0.1 & -1.41\% \\
    oxford & plants & 0.0001 & 0.0001 & -0.88\%\\
    food & fruit &  0.01 & 0.01 & 0.98\%\\
    fungus & plant & 0.01 & 0.01 & 0.78\%\\
    person & food & 0.01 & 0.01 & -0.71\%\\
    fruit & garment & 0.01 & 0.01 & -0.12\%\\
    music & plant & 0.01 & 0.01 & -2.86\%\\
    \hline
    \end{tabular}
\end{table}

\begin{figure}[ht]
   \includegraphics[width=\textwidth]{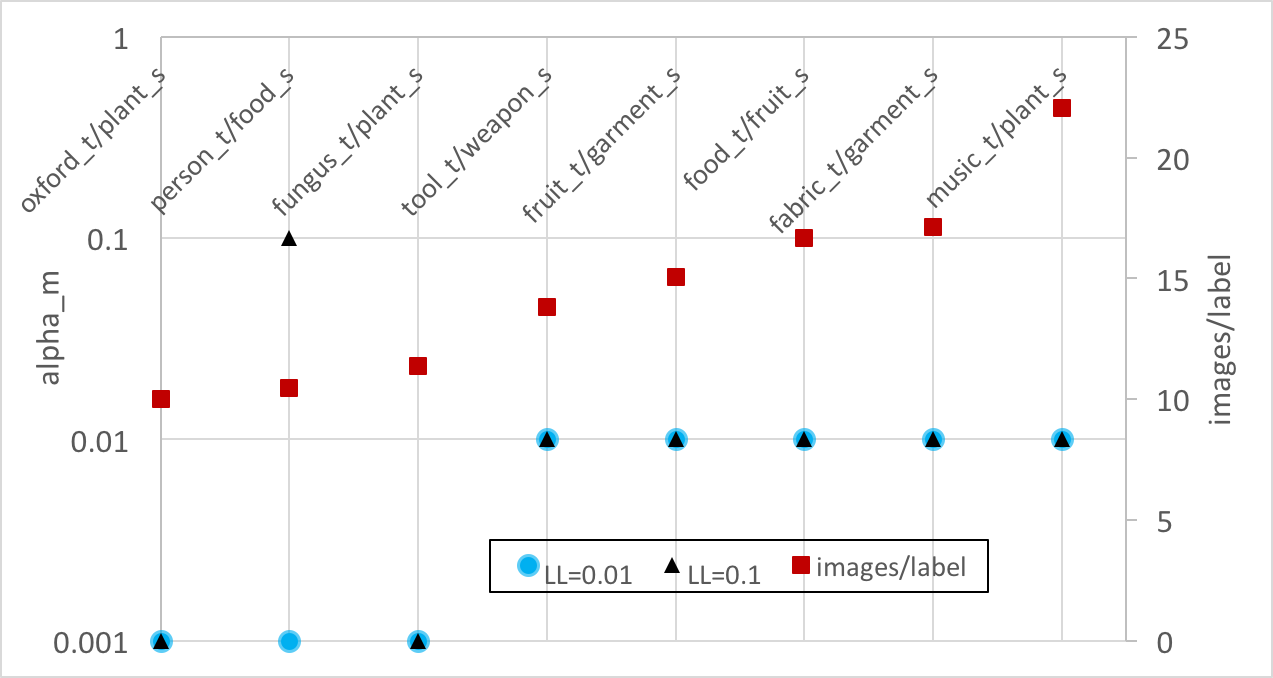}
   \caption{Correlation between $\alpha_m$ and images/label}
   \label{fig.alpham_img}
\end{figure}

We are interested in identifying correlation between source/target dataset features and $\alpha_m$. The first feature that we consider is images/label in the target dataset. Intuitively with more labelled data for the target domain, we can be more aggressive (i.e., use larger $IL$ and $LL$) in finetuning. Figure~\ref{fig.alpham_img} plots $\alpha_m$ versus images/label in target for $m=0.01$ and $0.1$. For both these cases we observe that $\alpha_m$ increases with images/label. However there is one anomaly, $\alpha_{0.1}=0.1$ for $person_t/food_s$, though $person_t$ has smaller images/label. This seems to allude that other features of source/target datasets also dictate the choice of learning rates. We are currently investigating this direction with the hope to develop some functional mapping between the features of source/target datasets and $\alpha_m$. This knowledge can be leveraged to develop intelligent algorithms to identify the best learning rate for inner layers and outer layers for a given source/target dataset. 

\subsection{Graduated Finetuning of Inner Layers}
We also investigated how the top-1 accuracy varies if the inner layer learning rate multipliers are  not kept at a fixed value but varied. With the assumption that very basic concepts learned in the earlier layers are more important for transfer learning  than later layers which map to complex concepts,  we varied the learning rate multipliers in steps within the inner layers.

\subsubsection{Oxford Flowers Dataset} 
\begin{figure}[ht]
   \includegraphics[width=\textwidth]{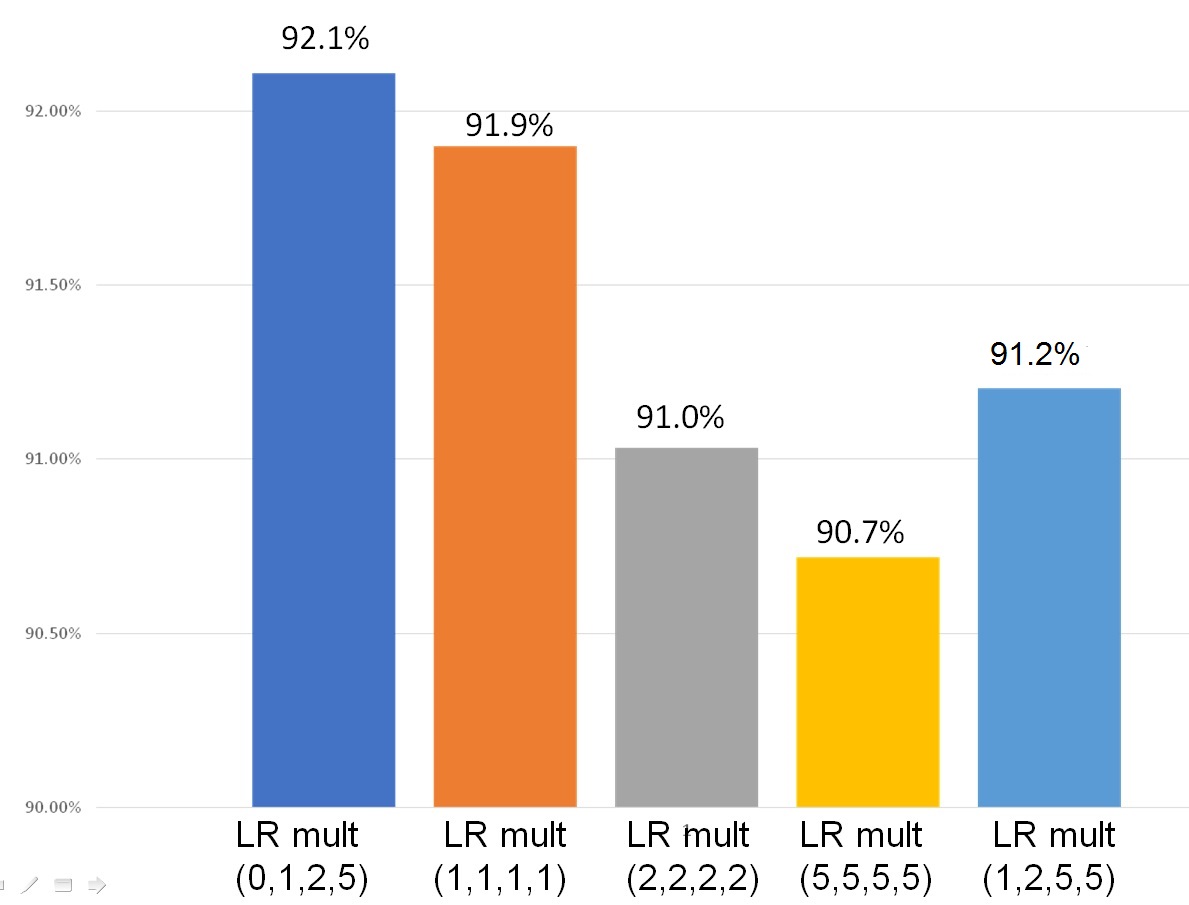}
   \caption{Top-1 accuracy with varying inner LR mult and fixed outer LR mult at 20}
   \label{fig.varying_lrmul}
\end{figure}

The ResNet-27 we are using for throughout these experiments has inner convolutional layers organized in 5 stages, conv1 through conv5  as shown in Figure \ref {fig.ResNet27}.  We can denote the learning rate multiplier for each of these 5 stages as $IL_1$ through $IL_5$.  We  measured the accuracy of finetuning when we kept the inner learning rate multiplier ($IL_1$..$IL_5$) equal across stages, (at a fixed value of either  1, 2 or 5) and also compared to using a graduated set of values.  In this case, each convolutional stage was assigned a multiplier (like 0, 1, 2, and 5), with conv1 and conv2 using the same (first, smallest) multiplier, and conv3, 4, and 5 using the successive, larger multipliers. (Meaning $IL_1$ was equal to $IL_2$.) In each case we set the learning rate multiplier $LL$ of the last layer to 10.  Figure \ref {fig.varying_lrmul} shows the top-1 accuracy for different $IL$ configurations with Oxford flowers as the target dataset and plant as the source data set with the base learning rate at 0.001. As the chart shows,  the best accuracy was achieved when the learning rate multipliers were graduated. 

 \begin{figure}[ht]
   \includegraphics[width=\textwidth]{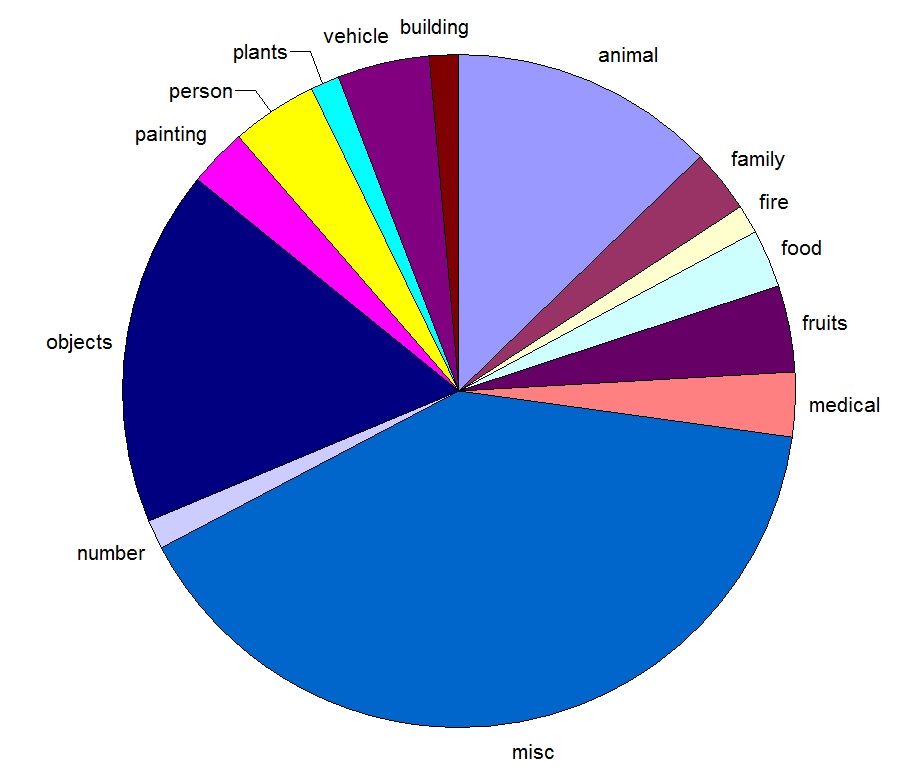}
   \caption{Distribution of Image Classification Tasks from service API used}
   \label{fig.custom1}
\end{figure}

\subsubsection{Real World Image Classification Tasks}
Next, we sought to validate these observations on training data "in the wild".  IBM operates a public cloud API called Watson Visual Recognition\footnote{https://www.ibm.com/watson/developercloud/visual-recognition/api/v3} which enables users to train their own image classifier by providing labelled example images.  While images provided to the API are not used to train anything aside from that user's model, users can opt-in to allow their image data to be used to help evaluate changes in the training engine.  From the many training tasks that were opted-in, we took a random sample of 70 tasks.  We did not manually inspect the images, but based on the names given to the labels, we presumed they represented a wide variety of image types, including industrial, consumer, scientific, and social domains as shown in  Figure \ref {fig.custom1}. Based on the languages of the class labels, we had a wide geographic range as well.  The average number of training images per task was about 250, with an average of 5 classes in each, so a mean of 50 image examples per class.  We randomly split these into 80\% for training and 20\% for validation, leaving 40 training images per class on average.

For each of the 70 training tasks, we created a baseline model that was a ResNet-27 initialized with weights from an ImageNet1K model.  We set the base learning rate to 0.001 and the $LL$ to 10.  The $IL$ was set to 0.  We fine-tuned the network for 20 epochs and computed top-1 accuracy on the held-out 20\% of labelled data from each task.  The average top-1 accuracy across the 70 tasks was 78.1\%.

For the graduated $IL$ condition, we initialized $IL_1$..$IL_5$ to be $\{0,1,2,4,8\}$ and $LL$ to be 16. We then defined a set of 11 scales, $\{0.25, 0.5, 1.0, 1.5, 2, 2.5, 3, 4, 5, 7, 10\}$.  The scale is a secondary learning rate multiplier.  For example, the final learning rate at scale 0.5 for conv3 ($IL_3$) and base learning rate 0.001 would be $0.5 * 2 * 0.001 = 0.001$.  The intuition is to combine the scale factors explored in Figures~\ref{fig.setAaccuracyVsIL} and \ref{fig.setBaccuracyVsIL} with the graduated values of $IL_1$..$IL_5$ explored in figure \ref{fig.varying_lrmul}.

This combination of scales and learning tasks resulted in $70 * 11 = 770$ additional finetuning jobs, which we ran for 20 epochs each.  We evaluated the top-1 accuracy for each of these jobs.  We found that if we picked the individual scale which maximized the accuracy for each job, the mean top-1 accuracy across all tasks improved from 78.1\% to 88.0\%, a significant gain.  However, to find this maximum exhaustively requires running 11 fine-tuning jobs for each learning task.  So we looked at which scale was most frequently the optimal one, and it was scale of 0.25.  If we limit ourselves to one finetuning job per training task, and always chose this single scale, the mean top-1 accuracy across jobs had a more modest increase, from 78.1\% to 79.7\%.

This promising direction needs further investigation; if we could predict the optimal learning rate multiplier scale based on some known characteristic of the training task, such as number of images per class, or total number of training images, we could efficiently reach the higher accuracy point established by our exhaustive search.

\section{Conclusion}
Transfer Learning is a powerful method of learning from small datasets.  However the accuracy obtained from this method could vary substantially depending on the choice of the hyperparameters for training as well as the selection of the source dataset and model. We study the impact of the learning rate and multiplier  which can be set for every layer of the neural network. We present experimental analysis based on the large ImageNet22K dataset, the small Oxford flower dataset and real world image classification datsets and show that the images per label parameter could be used to determine what the learning rates. It also seems like continuously varying the learning rate for inner layers has more promise than keeping them all fixed and is a worthy direction to pursue.

\bibliographystyle{splncs04}
\bibliography{main-davu18}

\end{document}